\pdfoutput=1

\documentclass[11pt]{article}

\usepackage{emnlp2021}
  
\usepackage{times}
\usepackage{latexsym}
\usepackage{subcaption}

\usepackage[T1]{fontenc}
 
\usepackage[utf8]{inputenc}

\usepackage{microtype}
\usepackage{booktabs}
\usepackage{graphicx}
\usepackage{xcolor}
\usepackage{amsfonts}       
\usepackage{nicefrac}       
\usepackage{microtype}    
\usepackage{hyperref}       
\usepackage{url}            
\usepackage{amsmath}
\usepackage{float}
\usepackage{comment}
\title{Parameter-Efficient Tuning Makes a Good Classification Head}

\author{Zhuoyi Yang\textsuperscript{\textdagger}\thanks{\ \ Equal contribution. Codes are at \url{https://github.com/THUDM/Efficient-Head-Finetuning}.}\quad Ming Ding\textsuperscript{\textdagger}\footnotemark[1]\quad Yanhui Guo\textsuperscript{$\ddagger$}\quad Qingsong Lv\textsuperscript{\textdagger}\quad Jie Tang\textsuperscript{\textdagger} \\
  \textsuperscript{\textdagger}Tsinghua University\quad \textsuperscript{$\ddagger$}Shandong Woman University \\
  \texttt{\{yangzy22,dm18\}@mails.tsinghua.edu.cn} \\
  \texttt{jietang@tsinghua.edu.cn} \\
  }

\begin{document}
\maketitle
\begin{abstract}
In recent years, pretrained models revolutionized the paradigm of natural language understanding (NLU), where we append a randomly initialized \emph{classification head} after the pretrained backbone, e.g. BERT, and finetune the whole model. As the pretrained backbone makes a major contribution to the improvement, we naturally expect a good pretrained classification head can also benefit the training. However, the final-layer output of the backbone, i.e. the input of the classification head, will change greatly during finetuning, making the usual head-only pretraining (LP-FT) ineffective. In this paper, we find that parameter-efficient tuning makes a good classification head, with which we can simply replace the randomly initialized heads for a stable performance gain. Our experiments demonstrate that the classification head jointly pretrained with parameter-efficient tuning consistently improves the performance on 9 tasks in GLUE and SuperGLUE.
\end{abstract}
\section{Introduction}
Fine-tuning is the most prevalent paradigm to leverage pretrained language models for the best performance on specific tasks~\cite{kenton2019bert}. Usually, a task-oriented classification head is grafted onto the final layer of the pretrained backbone, mostly Transformers~
\cite{vaswani2017attention}, and then the whole model is trained for the downstream task. Compared with training a large Transformer from scratch, the pretrained backbone has already learned to extract features about grammar, semantics and high-level understanding, making it easy to adapt the model for NLP tasks. 

Although the good initial weights of the pretrained backbone are the key factor of the effectiveness,
the initial weights of the classification heads are, however, largely under-explored. To the best of our knowledge, the usage of randomly initialized classification heads is still the overwhelmingly dominant choice in NLP.
Recently, LP-FT~\cite{kumar2022fine} finds that on many computer vision benchmarks, the performance can be promoted, especially for out-of-distribution (OOD) data, if we first only finetune the linear classification head (probe) with the pretrained backbone frozen, and then finetune the whole model. 

\begin{figure}[t]
    \centering
    \includegraphics[width=\linewidth]{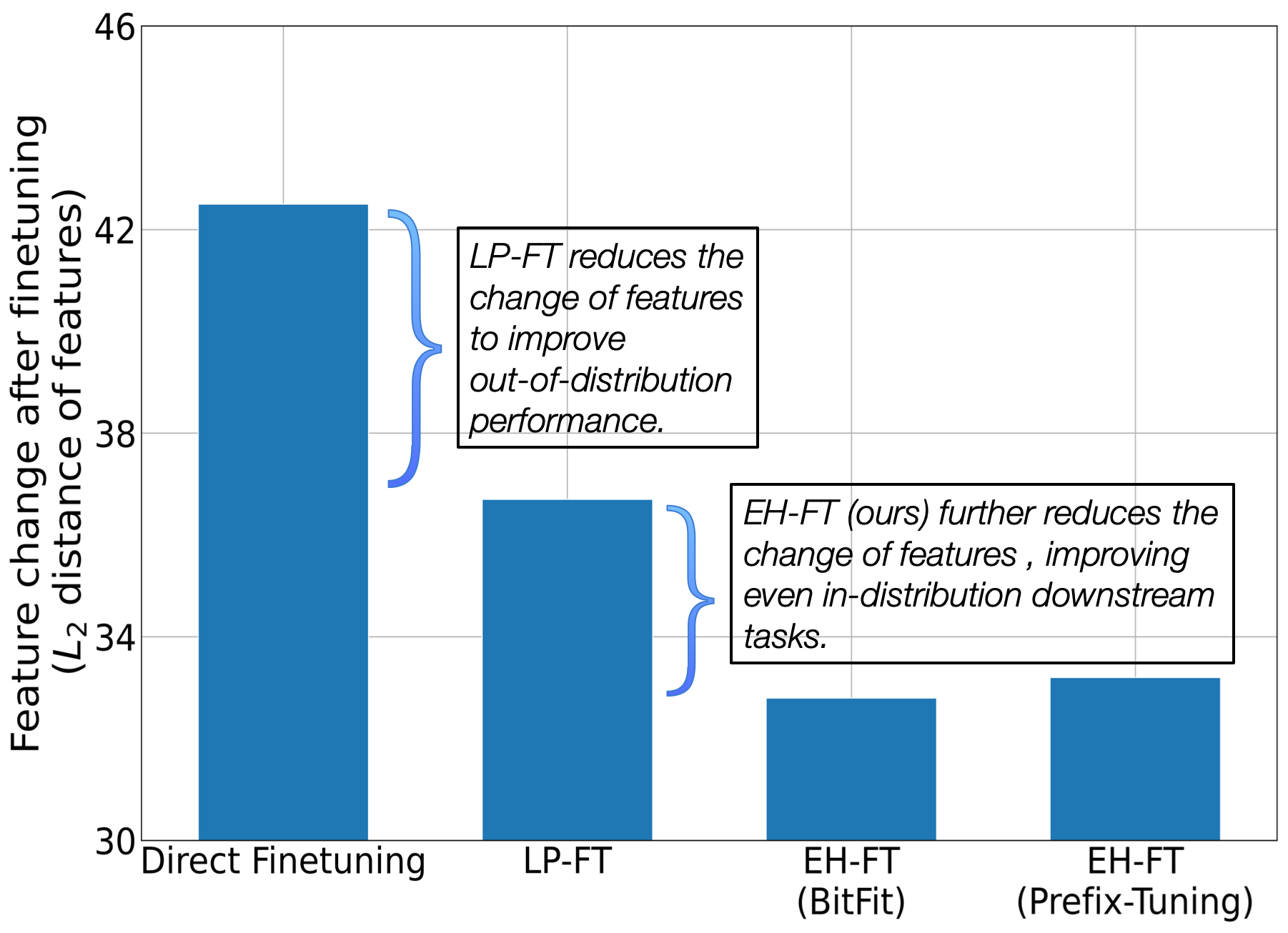}
    \caption{
        The amount of feature change after finetuning RoBERTa-large
        on RTE with different strategies, where ``features'' denote the final-layer output of the pretrained RoBERTa backbone, a.k.a. the input of the classification head. The feature change is measured by the $L_2$ distance between the features on the training set before and after finetuning.
    }\label{fig:feat_change}
    \vspace*{-5mm}
\end{figure} 

Does this kind of head-first finetuning technique also work in NLP?
Furthermore, since the principle is basically to reduce the change of features during finetuning (Figure~\ref{fig:feat_change}), is it possible to upgrade LP-FT to a general method to improve generalization, instead of only OOD setting? 
In this paper, we give a positive answer via parameter-efficient tuning. 

\begin{figure*}[t]
    \centering
    \includegraphics[width=\linewidth]{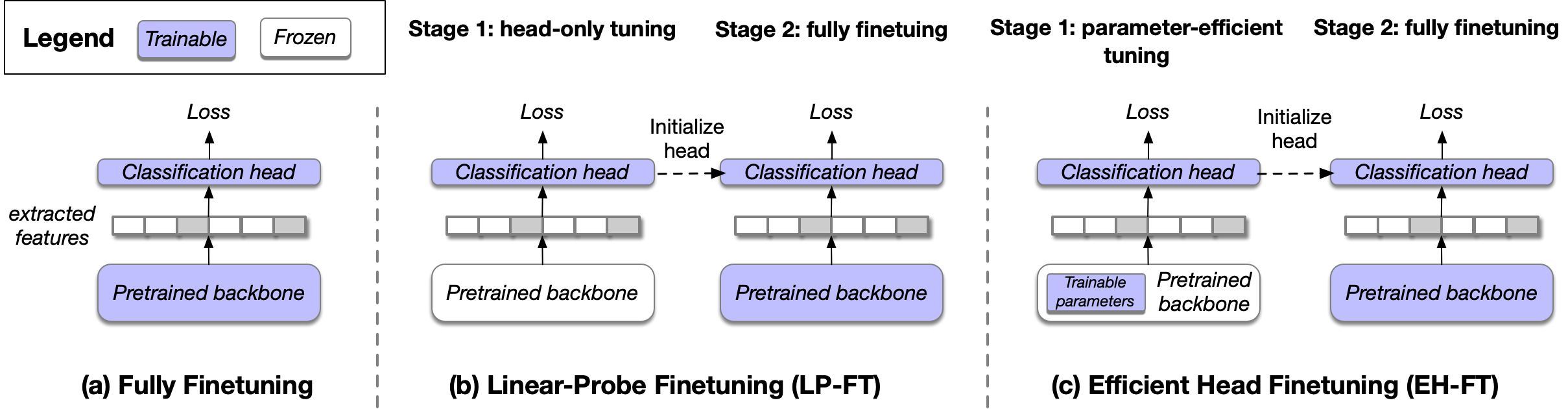}
    \caption{Illustration for different finetuning strategies. (a) Fully finetuning directly optimize all the parameters. (b) LP-FT first only trains the head then the whole model. (c) Our EH-FT first trains the head and a very small part (1\%$\sim$ 4\%) of parameters by a parameter-efficient tuning algorithm at the stage 1, and then restart a fully finetuning with the trained head from the stage 1.}
    \label{fig:ehtuning}
  \end{figure*}
To answer these questions, we need to first understand why a well-initialized head helps. LP-FT explains that with a randomly initialized head, the extracted features from the backbone change greatly during fitting the in-distribution (ID) samples in the training set, but the extracted features for OOD samples change very little. Finally, the training makes the features inconsistent for ID and OOD samples. A good classification head from linear probing can reduce the intensity of change in the extracted features during finetuning, and thus keep the feature consistency.

Since the overfitting-like explanation above only relates to the training samples, it should still be reasonable if we replace the ``ID/OOD samples'' by ``training/test samples'' in the explanation.
However, both our and the original LP-FT experiments suggest no significant improvement for the (ID) downstream task itself, why is that? 

In our opinion, the inconsistency is attributed to that \textbf{the head-only tuning in LP-FT cannot obtain a good enough classification head if the pretrained task is greatly different from the downstream task}. 
The head-only tuning usually has much worse performance than fully funetuning, so that even with the pretrained head, in the fully finetuning stage the weights in the backbone still need to change greatly to a totally different local minima for better performance. 

In this paper, we define the criterion of \emph{good classification heads} as those with which the pretrained backbone can be optimized to a near-optimal point with little change. The recently rising parameter-efficient tuning methods~\cite{li2021prefix, liu2021gpt,guo2021parameter,zaken2021bitfit,hu2021lora} aim to approach the performance of fully finetuning by only changing a very small part of parameters, and thus fulfil our desires.

Our experiments show that the parameter-efficient tuning methods, such as Prefix-tuning~\cite{li2021prefix}, BitFit~\cite{zaken2021bitfit} and LoRA~\cite{hu2021lora} indeed make good classification heads. Initialized by these good classification heads, the performances on 9 tasks of GLUE~\cite{wang2018glue} and SuperGLUE~\cite{wang2019superglue} are consistently better than or equal to that of direct finetuning or LP-FT. 

\section{Methodology}
\subsection{Background}
Consider a pretrained language model backbone $f(x;\theta_f)$, where $x$ is the input text and $\theta_f$ is the model parameters. $f$ servers as a \emph{feature extractor}. For a specific task, a learnable classification head $g(f(x;\theta_f);\theta_g)$ takes the extracted features as input and makes prediction for the downstream tasks. In previous research, three methods to train the models are widely used:

\paragraph{Finetuning.}
In traditional fully fine-tuning, $\theta_g$ is randomly initialized and trained simultaneously with the pretrained backbone $\theta_f$. 
\paragraph{Linear Probing.} Linear probing refers to head-only tuning, mainly used for evaluating the self-supervised learning representations in computer vision. The classification head $\theta_g$ is randomly initialized and trainable, but the backbone $\theta_f$ is frozen during training.
\paragraph{Linear-probing finetuning (LP-FT).} LP-FT is two-stage tuning method recently proposed by \citet{kumar2022fine}. The stage 1 of LP-FT is linear probing, and the stage 2 is fully finetuning with the classification head initialized as the trained head in the stage 1. This method proves to be better than finetuning or linear probing for OOD samples.
\paragraph{Parameter-efficient Tuning.} Recently, a collection of new tuning methods for pretrained models aims to approach the finetuning performance by only changing a small part of parameters, which is called parameter-efficient tuning. It includes methods by limiting the trainable parameters, e.g. BitFit~\cite{zaken2021bitfit}, and methods by adding a small trainable module, e.g. LoRA~\cite{hu2021lora} and Prefix-Tuning~\cite{li2021prefix}. 

\subsection{Efficient Head Finetuning}
We propose the Efficient Head Finetuning (EH-FT) to step further about the finetuning paradigm, which can improve the downstream task in a simple way. 

EH-FT is also a two-stage method similar to LP-FT, but replacing the linear-probing in the first stage as parameter-efficient tuning. Figure~\ref{fig:ehtuning} illustrates the difference between these methods. Specifically, the procedure of EH-FT can be described as follows:

\paragraph{Stage 1.} We finetune the model consisting of the pretrained backbone $\theta_f^0$ and a randomly initialized classification head $\theta_g^0$ using a parameter-efficient tuning algorithm. At the end of training, we restore the pretrained backbone back to $\theta_f^0$ and only keep the trained head $\theta_g^*$. 
\paragraph{Stage 2.} We fully finetune the model consisting of the pretrained backbone $\theta_f^0$ and the classification head $\theta_g^*$ from the first stage. 

\subsection{Working Principle}
\label{section:work}
\begin{figure}[t]
  \centering
  \includegraphics[width=\linewidth]{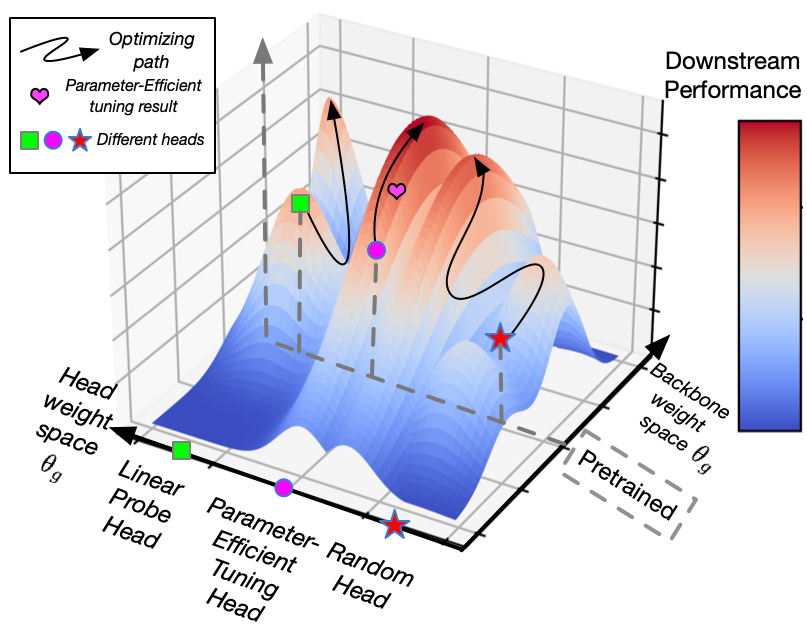}
  \caption{Diagram to show how different heads affect the optimizing path during finetuning. (a) Random head makes the early gradients very stochastic and largely changes the weights of the backbone. (b) Linear probing head is well adapted to the pretrained backbone, but the backbone weights still need to change a lot to provide specific information to reach a higher downstream performance. (c) Efficient head ensures the existence of a nearby good weight for the backbone (the result of the parameter-efficient tuning), the landscape helps quickly find a nearby optimum without largely changing the features. }
  \label{fig:optim}
\end{figure}

In this section, we will detail the working principle via comparing the optimizing path during finetuning with different initial heads, which is illustrated in Figure~\ref{fig:optim}. 

\paragraph{The initialization of classification head matters.}
As observed by~\citet{dodge2020fine} via varying random seeds, some specific heads are consistently significantly better than others on a group of binary classification tasks.
Our explanation is that if the pretrained backbone cannot be quickly optimized to a nearby local optimum, the features will change a lot, and thus affect the performance according to \citet{kumar2022fine}. However, most randomly initialized heads will back-propagate very chaotic gradients to the backbone, causing a large feature change and finally catastrophic forgetting and overfitting.

Therefore, we need a way to stably find a good pretrained classification head for finetuning, which is also the motivation of LP-FT.


\paragraph{LP-FT neglects the difference between the pretraining and downstream task. }The experiments in the paper of LP-FT~\cite{kumar2022fine} are mainly about finetuning a \emph{contrastive} pretrained backbone, e.g. CLIP~\cite{radford2021learning}, for classification tasks. However, when we apply \emph{generative} pretrained backbones, e.g. BERT~\cite{kenton2019bert} and MAE~\cite{he2022masked}, to classification tasks in NLP, the linear probing performs poorly because the extracted features are specialized for mask prediction and might not contain enough information for classification~\cite{wallat2020bertnesia, yosinski2014transferable}. Even though we can find the best classification head w.r.t. the pretrained backbone in the stage 1, the features are learnt to change greatly to adapt for the downstream task, which goes against the theory of LP-FT.

\paragraph{Efficient head ensures a nearby near-optimal point for finetuing the backbone weights.}
Although parameter-efficient tuning is originally proposed to reduce memory usage, a recent study~\cite{vu2022overcoming} finds that these methods restrict the trainability of parameters and thus helps overcome catastrophic forgetting. Our experiments also show that \emph{most parameter-efficient tuning methods change the feature much less than finetuning}, e.g. 20.0 for Prefix-Tuing verus 42.5 for finetuning measured in the $L_2$ distance in Figure~\ref{fig:feat_change}.

This property means that it could be easy to reach around the parameter-efficient tuning weights during a fully finetuning, and parameter-efficient tuning usually performs near optimally and much better than linear probing. 

Here we can hypothesize that EH-FT works in a way as follows:
\begin{itemize}
  \item In stage 1, parameter-efficient tuning acts like a weak surrogate of the fully finetuning on the backbone, and learns to add the most important information to the features for the downstream task. The efficient head adapts to these slightly changed features.
  \item In stage 2, the efficient head will guide the backbone quickly towards the region around the parameter-efficient tuning result by backpagation, because the region only differs in a small part of parameters and very near, and thus finally converges to a near local optimum. 
  
\end{itemize}
As a proof, The $L_2$ distance (as measured in the in Figure~\ref{fig:feat_change}) between EH-FT (Prefix-Tuing) Stage 1 and Stage 2 is only 27.4~\footnote{Similarly, 27.9 for EH-FT (BitFit).}, less than the 36.7 of LP-FT. EH-FT indeed converges to a local optimum near the parameter-efficient tuning result. 



\section{Experiments}

\begin{table*}[t] 
\centering
\setlength{\tabcolsep}{1.6mm}
\small{
\begin{tabular}{ccccccccccc} 
\toprule 
Methods & RTE & BoolQ & COPA & CB & WIC & MRPC & QNLI & COLA & STS-B & \textbf{Avg}  \\
\midrule 
Finetuning~\cite{liu2019roberta} &	86.60\ddag&	86.90&	94.00&	98.20\ddag&	75.60&	90.90\ddag&	94.70&	68.00 & 92.40\ddag& 87.48 \\
Finetuning (reproduce) & 87.52 & 86.32 & 93.75 & 94.64 & 73.90 & 90.81 & 94.75 & 68.72  & 92.27 &86.96 \\ 
\midrule 
Linear Probing & 61.10 & 64.31 & 80.25 & 79.47 & 67.97 & 75.45 & 69.95 & 33.17 & 60.25 & 65.77 \\ 
Prefix-Tuning   & 77.00&	83.90&	87.75&	\textbf{100}&	65.05&	89.56&	94.55&	57.89& 89.92 &82.85 \\ 
LoRA  & 88.50 & 86.29 & 94.75 & \textbf{100} & 73.04  & 90.43 & 94.37 & 67.86  & 92.17 &87.49\\ 
BitFit & 87.05&	86.13&	95.50 &	99.11&	72.01&	90.32&	94.68& 57.89 & 92.05&87.24\\ 
\midrule 
Top-K Tuning  & 86.55 & 85.34 & 93.00 & 98.66 & 73.71 & 90.94 & 94.12 & 65.78 & 91.47 & 86.62 \\
Mixout  & 85.56 & 86.06 & 95.00 & 98.66 & 74.45 & 90.87 &94.18 & 65.45 & 91.62 & 86.87 \\
Child-Tuning$_D$  & 88.18 & 86.65 & 94.5 & 92.86& 74.07 & 91.36 &94.44 & 68.52 & \textbf{92.51} & 87.00 \\
LP-FT  & 86.14 & 86.38 & 94.00 & 93.50 & 74.73 & \textbf{91.47} & 94.78 & 67.45  & 92.20 & 86.74 \\ 
\midrule 
EH-FT$_{\text{LoRA}}$ & \textbf{88.68} & 86.69 & 94.5 & 99.12 & 74.65 & 91.00 & 94.73 & 69.00  & 92.24&\textbf{87.85}\\ 
EH-FT$_{\text{PT}}$ & 87.22 & 86.9 & \textbf{95.00} & \textbf{100} & 73.63 & 90.56 & \textbf{94.89} & \textbf{69.10} & 92.31& 87.73\\ 
EH-FT$_{\text{BitFit}}$ & 88.10 & \textbf{86.97} & 94.75 & 99.12 & \textbf{75.20} & 91.00 & 94.61 & 68.78  & 92.25 & \textbf{87.86}\\ 
\bottomrule 
\end{tabular}
}
\caption{Results with RoBERTa-Large. All scores are the mean result of 4 random seeds. Results with \ddag\ finetuned starting from the MNLI model, which is expected a better performance than single-task finetuning. It is recommended to compare the reproduced finetuing result with LP-FT and EH-FT, because they share the same code, only with different head initialization.
}
\label{2STEPresult1}
\end{table*}

\begin{table*}
\centering
\small{
\begin{tabular}{ccccccccccc} 
\toprule 
Methods & RTE & BoolQ & COPA & CB & WIC & MRPC & QNLI & COLA & STS-B & \textbf{Avg}  \\
\midrule 
Finetuning & 76.08 & 79.79 & 75.50 &	91.52 &	\textbf{71.71} &	90.90 &	\textbf{92.54} &	63.44 &	90.60 & 81.34 \\

EH-FT$_{\text{BitFit}}$ & \textbf{76.35} & \textbf{80.95} & \textbf{77.00} & \textbf{97.77} & 71.52 & \textbf{92.10} & 92.34 & \textbf{64.91}  & \textbf{90.64} & \textbf{82.62}\\ 
\bottomrule 
\end{tabular}
}
\caption{Results in BERT-Large.  All scores are the mean result of 4 random seeds. EH-FT$_\text{BitFit}$ outperforms fine-tuning in 7 of 9 tasks.
}
\label{bertResult}
\end{table*}

\subsection{Datasets}
SuperGLUE  \citep{wang2019superglue} is a benchmark which contains 8 difficult natural language understanding tasks, including BoolQ, CB, COPA, MultiRC, ReCoRD, RTE, WiC, and WSC. In our experiments, we exclude WSC, ReCoRD, MultiRC because they rely on a heavy pipeline instead of a single classification head on RoBERTa-Large to get a satisfying result. We further supply 4 widely used GLUE datasets, MRPC, COLA, QNLI and STS-B into our benchmark.
We report our results on the dev set following most previous works~ \cite{liu-etal-2022-p,xu2021raise}.


For COLA, we evaluate performance using matthews correlation coefficient. For MRPC, we use F1 score.  For STS-B, we use Pearson correlation coefficients with RoBERTa-Large and Spearman correlation coefficients with BERT-Large in order to be consistent with the baseline  \citep{liu2019roberta, kenton2019bert}. We use accuracy in other tasks.

\subsection{Setup}

\paragraph{Models} We mainly use RoBERTa-Large~ \cite{liu2019roberta} as the pretrained backbone in our experiments, and mostly follow the best hyperparameter settings from the original paper. To exhibit the generality of EH-FT, we also conduct some experiments on BERT-Large. The pretrained weights are obtained from HugggingFace \cite{wolf2019huggingface}, and the codes are bases on the SwissArmyTransformer~\footnote{\url{https://github.com/THUDM/SwissArmyTransformer}} framework. 
We set the classification head as a 2-layer MLP mapping from the hidden dimension of the model to 2,048, and then to the number of classes. 
Experiments are executed on DeepSpeed library and NVIDIA A100 GPUs with mixed precision floating point arithmetic. We report the average results over 4 random seeds. We implement BitFit, LoRA and Prefix-Tuning as the parameter-efficient module in Stage 1.

\paragraph{Hyperparameters} 
Learning rate plays an important role in model training. However, the best learning rates on various datasets tend to be consistent. we fix the learning rate to 1e-5 for RoBERTa-Large (except for the WIC where we use 3e-5), and 3e-5 for Bert-Large. We set the batch size=32, and use AdamW~\cite{loshchilov2017decoupled} optimizer with $\beta_1$=0.9, $\beta_2$=0.98, $\epsilon$=1e-6, weight decay=0.1. Following the original finetuing strategy of BERT, we adopt a warmup for the early 10\% iterations and then a linear learning rate decay. 
To fairly compare with two-stage method and the direct finetuning, we need to keep the same entire training iterations. This is implemented by recording the iterations to convergence for each dataset, and dividing the total number of iterations into 10\% and 90\% respectively for Stage 1 and Stage 2. 

We set learning rate to 5e-4 for BitFit, LoRA and Linear probing, 5e-3 for Prefix-Tuning. Prefix number is set to 16 for Prefix-Tuning and intermediate dimension $r$ is set to 32 for LoRA. The learning rates are determined by a grid search in \{5e-3, 1e-3, 1e-4, 5e-4\} on RTE and BoolQ.
 
\subsection{Other finetuning Strategies}
We also compare the EH-FT results with other finetuning strategies, but note that they should not be directly seen as baselines because most of them are compatible with EH-FT.

\paragraph{Top-K Tuning}
Top-K Tuning~\cite{yosinski2014transferable} finetunes only the top $k$ layers while freezing the others. This method can be considered  a strategy to prevent overfitting. Following the setting of \cite{xu2021raise}, we report the best value by varying the layer number $k$ from \{3, 6, 12\}. 

\paragraph{Mixout}
Mixout \cite{lee2020mixout} randomly replaces model parameters by pretrained weights with probability $p$ during finetuning in order to reduce the deviation of model parameters. Following this paper, we search the optimal $p$ from \{0.7,0.8,0.9\}, and learning rate from \{1e-5, 2e-4\}. 

\paragraph{Child-Tuning}
Child-Tuning \cite{xu2021raise} updates a subset of parameters (called child network) during the backward process. The subset is chosen randomly (Child-Tuning$_F$) or chosen with the lowest Fisher Information (Child-Tuning$_D$).
We search the optimal subset ratio $p$ from \{0.1, 0.2, 0.3\} and only implement Child-Tuning$_D$.  We conduct experiments based on their public code \footnote{\url{https://github.com/RunxinXu/ChildTuning}}. 

\subsection{Results}
\paragraph{RoBERTa-Large} We report the experimental results in Table \ref{2STEPresult1}. Besides fully finetune and EH-FT,  we also show the results of enhanced methods for finetuning and parameter-efficient tuning.  It can be seen that all three kinds of EH-FT outperform fully finetuning, providing the improvement of 0.9, 0.89 and 0.77 in average score. EH-FT works well on RTE, COPA, CB and WIC. In the other hand, on some very simple tasks (MRPC, COLA) or tasks with a large training set (QNLI), the overfitting and forgetting problem is not obvious and the performance is hard to improve with EH-FT.  

For LP-FT, there is no significant improvement in the in-distribution data, which is consistent with our speculation in section \ref{section:work}. Other improvement methods also have been found perform poorly on certain datasets. Using BitFit and LoRA in Stage 1 can achieve stable performance, but prefix-tuning shows high variance on different datasets. We guess that this may be the consequence of its instability  \citep{chen2022revisiting}. Moreover, our approach has the same advantage as parameter-efficient tuning, preventing the model from overfitting on low-resources datasets (CB), which brings a great improvement in performance. 

\begin{figure*}[h]
  \centering
  \includegraphics[width=1\linewidth]{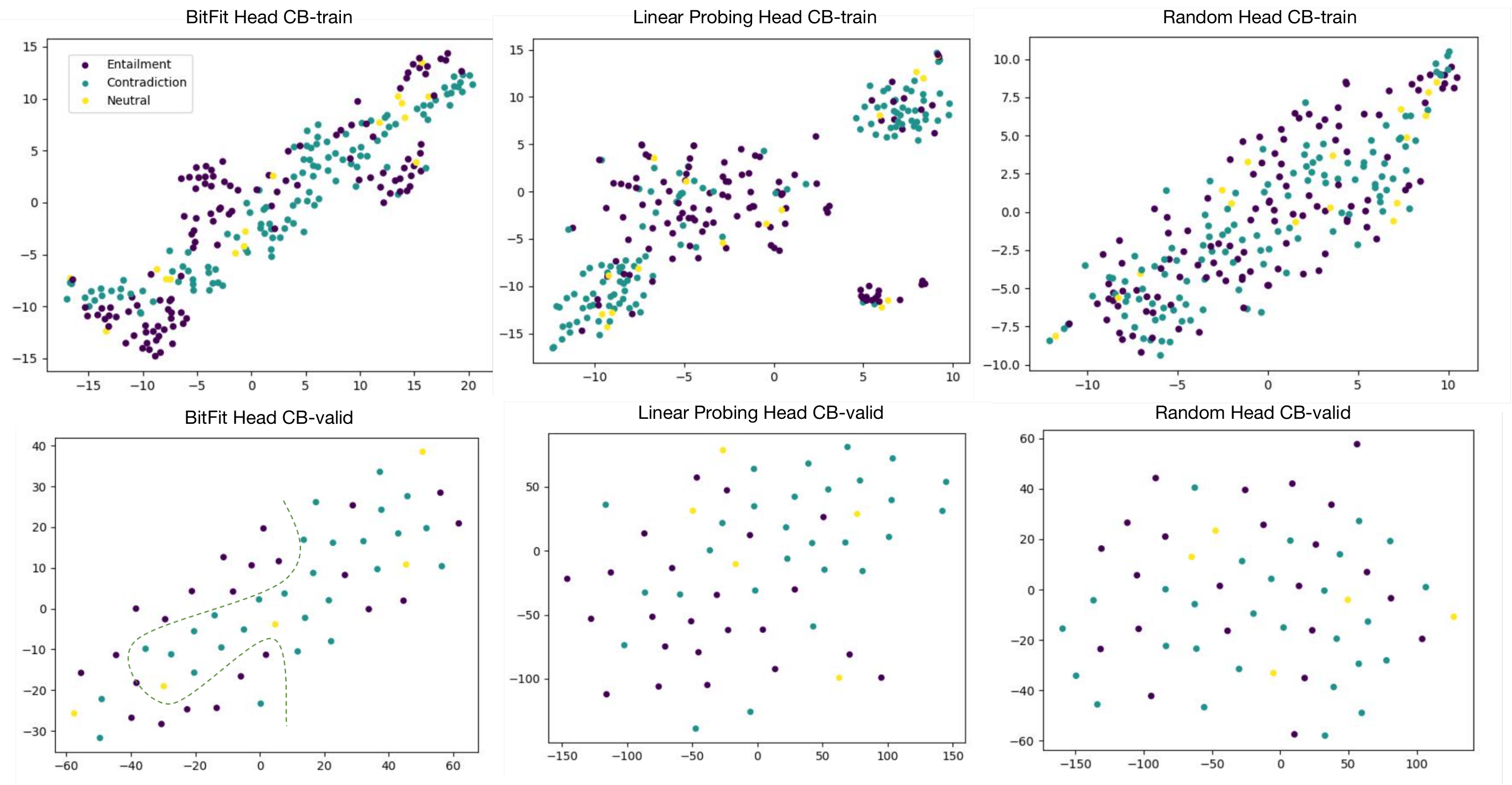}
  \caption{Features distribution with different head initialization strategies. We reduce the dimension of the features output by classification using T-SNE. The upper plots show the distribution of samples in CB training set, and the lower plots show the samples in development set. Different labels are distinguished by point color. \textbf{Parameter-efficient tuning can make a good clustering which is not highly concentrated.}}
  \label{embedding}
\end{figure*}
\paragraph{BERT-Large} We also do comparison experiments between fully finetune and EH-FT$_\text{BitFit}$ in BERT-Large. Results are shown in Table \ref{bertResult}. EH-FT yields improvement of up to 1.28 average score on BERT-Large.
\section{Analysis}

\subsection{From the view of continual learning}
\begin{figure}
  \centering
  \includegraphics[width=1\linewidth]{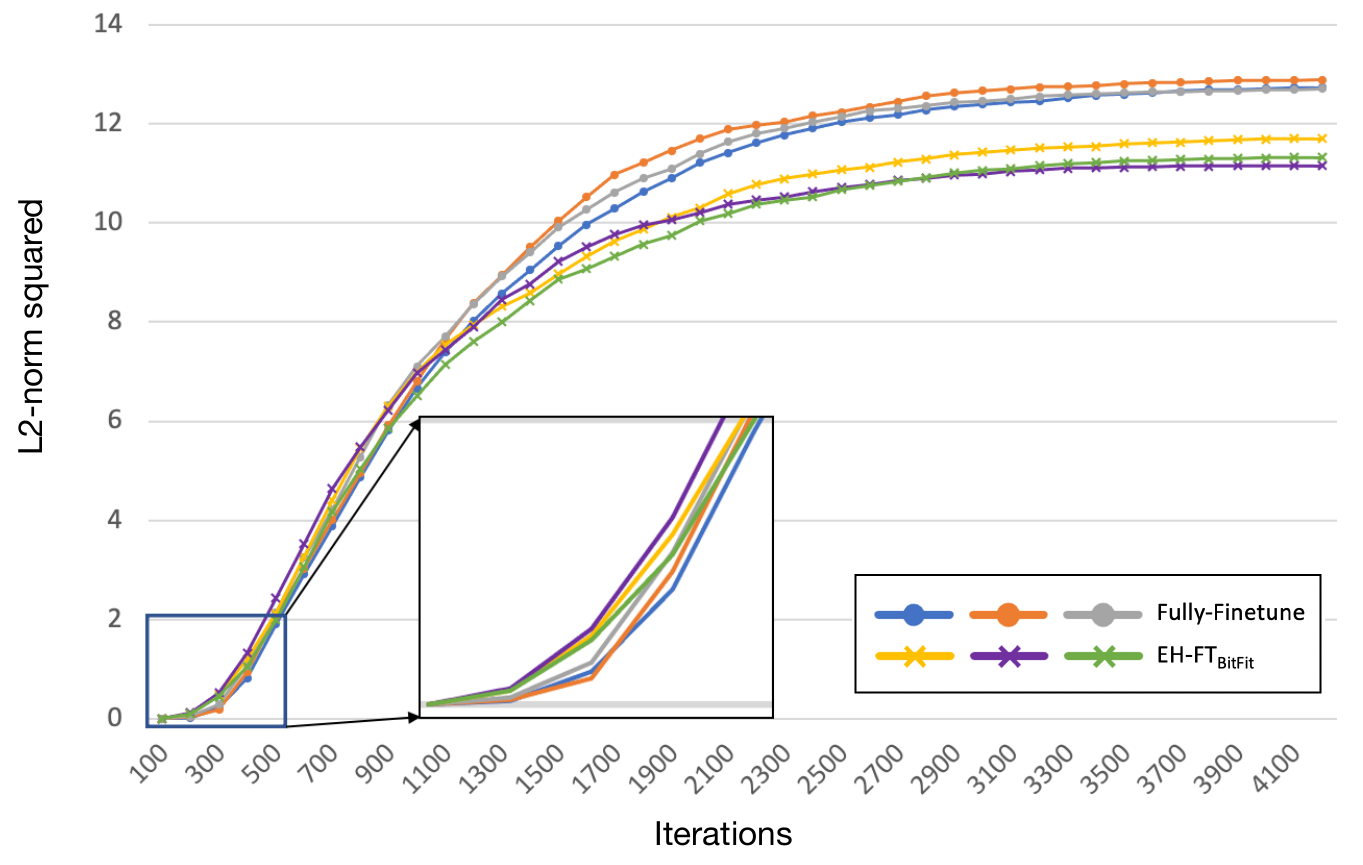}
  \caption{We present the $L_2$ distance between model weights and pretrained weight ($||\theta_f -\theta_f^*||^2$) on RTE dataset with multi runs. EH-FT will change parameters more quickly in the beginning but finally converge to a smaller distance. }
  \label{diff}
\end{figure}
To reduce the change of parameters during finetuning is a very common technique in continual learning~\cite{chen2020recall}. We find EH-FT has a similar effect.
We ran 3 random seeds under both fully finetuning and EH-FT$_{\text{BitFit}}$  (Stage 2) in RTE dataset and visualized the $||\theta_f -\theta_f^*||^2$ (Euclidean distance between finetuning parameters and pretrained parameters) in Figure \ref{diff}. 
The curve of EH-FT$_{\text{BitFit}}$ raises faster in the first 1,000 iterations than fully finetuning due to its large gradient in the beginning. But EH-FT$_{\text{BitFit}}$ converges to a small value than fully finetuning in the end.

\subsection{Better Initialized Distribution} 
We expect that the classification head obtained by parameter-efficient method can bring a better initialized distribution which can guide the model to converge fast with the prior knowledge learned in Stage 1. 

Considering classes\_number is usually small, we take the hidden states in the middle layer of the head, whose dimension is 2048 as features generated by head. We choose the CB dataset which has 250 train samples and 57 validation samples and reduces the dimension of feature using T-SNE. In Figure \ref{embedding}, the classification head initialized by BitFit has a good clustering which is easier for the model to fit. As a comparison, linear probing makes a more concentrated clustering but with low accuracy. This type of clustering may make it difficult for the model to correct it back in Stage 2. The features obtained by a random head are uniformly distributed over the entire space, with no clear dividing lines among the different categories. This may cause chaotic gradients in the early stage of fully finetuning and lead to knowledge forgetting.

\subsection{Time and Space Consumption}
Most of the current works to improve the finetuning performance in NLP introduce a non-negligible computational cost. Finetuning time would increase a lot because of the extra loss terms (RecAdam~\cite{chen2020recall}, SMART~{\cite{jiang-etal-2020-smart}}), extra modules (Mixout~\cite{lee2020mixout}) or gradient preprocessing (Child-Tuning~\cite{xu2021raise}). 

In addition, some of them also have large space consumption. RecAdam and Mixout need to keep a copy of the pretrained parameters and access them every training step, and Child-Tuning stores a boolean variable for every parameter. 

We show that EH-FT is significantly more computationally efficient and spatially efficient than the above methods without hurting performance.


\paragraph{Time Consumption} Our method introduces two additional parts of extra computation during training. The first part is the time spent by  additional parameter-efficient module added in Stage 1. However, most of those modules are computationally efficient, and BitFit does not even have extra computation. Furthermore, extra modules only exist in Stage 1 which has few epochs. Another is the overhead of restarting training in Stage 2. This part is negligible in most cases and can be also avoided by carefully programming. 

It is noteworthy that thanks to the rapidly converging during Stage 2, the total epochs (Stage 1 + Stage 2) can be kept in line with fully finetuning.

\paragraph{Space Consumption}
Just like the parameter-efficient methods above, EH-FT only needs 0.01\% to 3\% additional parameters. For BitFit, EH-FT needs to memorize the original bias and restore them before Stage 2. After training, saved checkpoint does not need to store those additional parameters.

Furthermore, it is easy to implement EH-FT in various deep learning frameworks: one just need to start the training twice. Since we do fully finetuning in Stage 2, it can be combined with any other methods.





\subsection{Ablation Study on EH-FT}
We study some factors that may affect the experimental performance, and compare the results varying the hyper-parameters.

\begin{figure}
  \centering
  \includegraphics[width=1\linewidth]{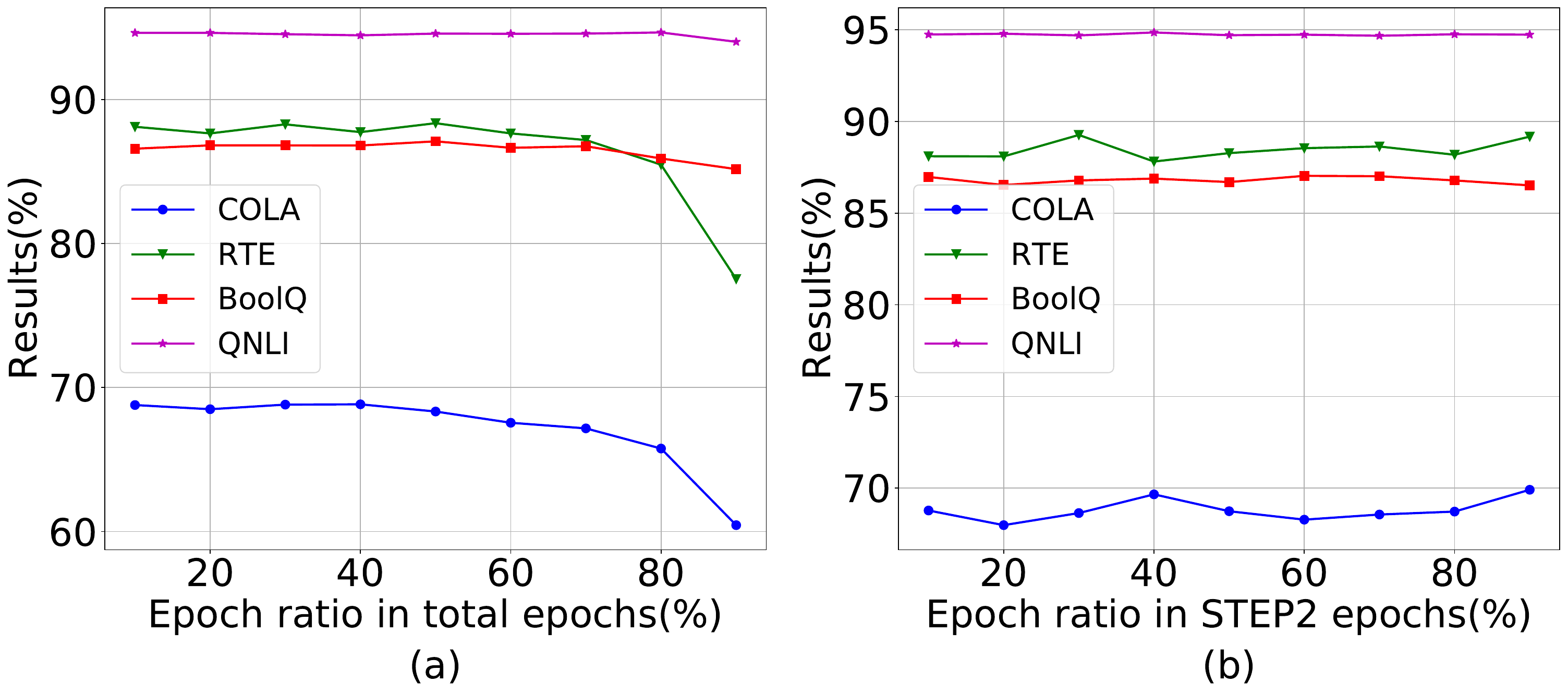}
    \caption{Ablation study on Stage 1 epochs of EH-FT$_\text{BitFit}$ using RoBERTa-Large. (a) Results with different proportions of Stage1 epochs while keeping the total epochs fixed. Training model for more epochs in Stage 1 can not substitute fully-finetune. (b) Results with different proportions of Stage 1 epochs while keeping the Stage 2 epochs fixed. Epochs of Stage 1 do not have a significant effect on the results. }
  \label{abation1}
\end{figure}

\paragraph{Effect of epoch proportion in the Stage 1. } We explore whether the training epochs of Stage 1 and Stage 2 will affect the performance when the total epochs remain the same. We increase the epoch proportion of Stage 1 from 10\% to 90\% and obtain the performance of RoBERTa-Large on CoLA, RTE, BoolQ and QNLI datasets. The results are shown in Figure \ref{abation1}. We found that when the proportion exceeded 50\%, there was a significant decrease in performance. This indicates that a sufficiently trained head cannot substitute the role of enough training in Stage 2.
In this paper, we uniformly use 10\% epochs as the first step of training for all the datasets, to ensure that parameters are fully trained. 

Besides, we also study the effect of increasing the training iterations of Stage 1 on the performance of the model when the training iterations of Stage 2 are fixed. We change the epoch proportion of Stage 1 from 10\% to 90\% and keep the Stage 2 epoch fixed. The results are also illustrated in Figure \ref{abation1}. It can be observed that doing parameter-efficient tuning for a long time does not affect the performance in Stage 2. 10\% epochs can  provide a good enough initialization of the classification head.

\begin{figure}
  \centering
  \subcaptionbox{\label{fig:subfig-a}}
    {\includegraphics[width=0.49\linewidth]{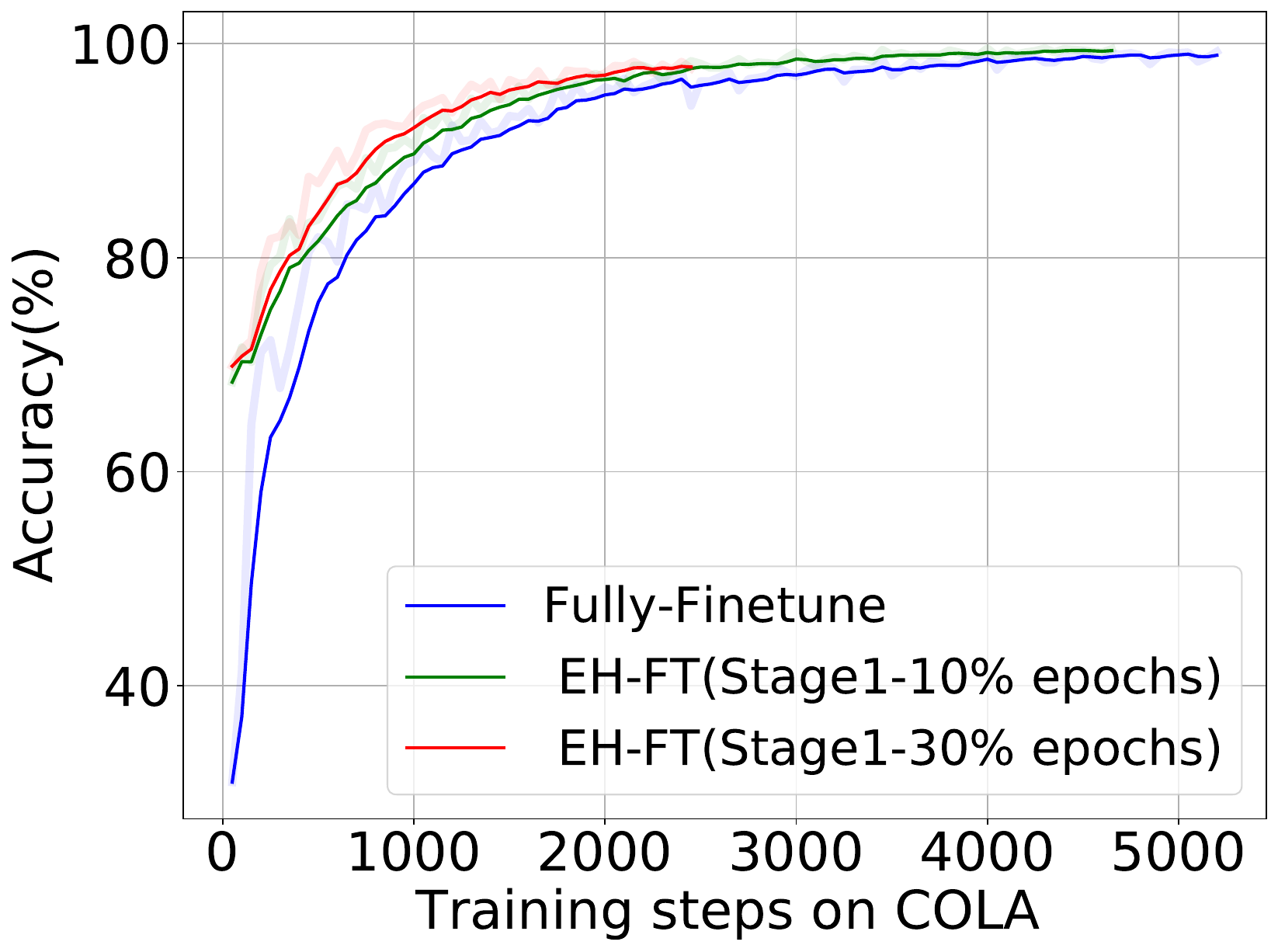}}
  \subcaptionbox{\label{fig:subfig-b}}
    {\includegraphics[width=0.49\linewidth]{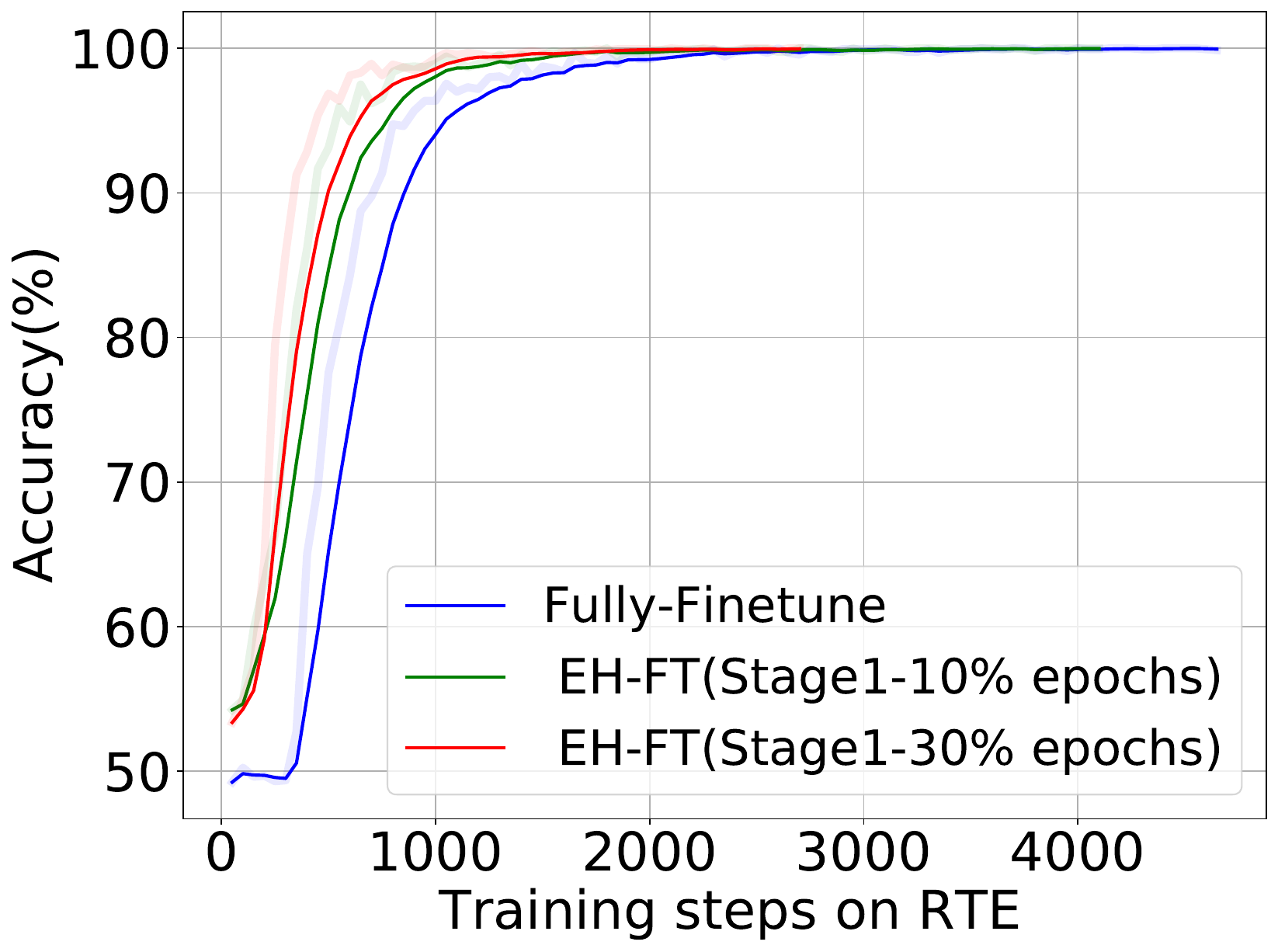}}
  \caption{Convergence rate with fully-finetune and EH-FT$_\text{BitFit}$ Stage 2. For EH-FT, we draw the curve with different Stage 1 epochs. Assisted by a good initialized head, the model can converge quickly in Stage 2. Increasing the training time of Stage 1 can increase the convergence rate of Stage 2. (a) On COLA training set. (b) On RTE training set. }
  \label{abation2}
\end{figure}

\noindent
\paragraph{Effect of epoch proportion on convergence of model. } 
To verify whether a good classification head can help learning and convergence of the model, we analyze the rate of convergence on the training set of COLA and RTE. We have drawn the result curves of three experiments: fully finetuning, EH-FT which allocate 10\% and 30\% of the total epochs to Stage 1, as shown in Figure \ref{abation2}. 
Because of the good initialized distribution brought by a good initialized head, EF-FT can converge more quickly than fully-finetune in Stage 2. But the convergence rate increases slightly when we adjust the Stage 1 ratio from 10\% to 30\%. 
That means if we do not run too many iterations in Stage 1, EH-FT (including the time for Stage 1) can converge almost at the same time as fully finetuning.

We also find that the classification head initialized by parameter-efficient tuning lead to larger gradients than the randomly initialized head by about two orders of magnitude. This may help the model to go directly to the period of rapid loss decrease.

\begin{table*}
\centering
\small{
\begin{tabular}{ccccccccccc} 
\toprule 
Methods & RTE & BoolQ & COPA & CB & WIC & MRPC & QNLI & COLA & STS-B & \textbf{Avg}  \\
\midrule 
Finetuning & 87.52 & 86.32 & 93.75 & 94.64 & 73.90 & 90.81 & 94.75 & 68.72 & 92.27 & 86.96 \\ 
EH-FT$_{\text{BitFit}}$ & 88.10 & 86.97 & 94.75 & 99.12 & 75.20 & 91.00 & 94.61 & 68.78 & 92.25 & \textbf{87.86} \\ 
EH-FT$_{\text{BitFit}}$-reserve & 87.37 & 86.40 & 95.50 & 99.55 & 75.00 & 90.32 & 94.54 & 69.31 & 92.27 & 87.80 \\ 
EH-FT$_{\text{LoRA}}$ & 88.68 & 86.69 & 94.5 & 99.12 & 74.65 & 91.00 & 94.73 & 69.00 & 92.24 & \textbf{87.85} \\ 
EH-FT$_{\text{LoRA}}$-reserve & 87.37 & 80.38 & 92.22 & 99.55 & 75.01 & 90.38 & 94.62 & 68.57 & 92.22 & 87.12 \\ 
\bottomrule 
\end{tabular}
}
\caption{We reserve the parameters tuned on Stage 1 in EH-FT(BitFit)-reserve and EH-FT(LoRA)-reserve. There is a slight decrease in overall results.
}
\label{ab1}
\end{table*}

\noindent
\paragraph{What if we don't remove the additional parameters before Stage 2?}
We reserve the parameters tuned in Stage 1 and conduct experiments on all the above datasets. Results are shown in table \ref{ab1}. The EH-FT-reserve performs slightly worse than EH-FT but still better than finetuning. The loss and accuracy of EH-FT-reserve hardly change in Stage 2, which indicates it is likely to be trapped in a local minimum and hard to optimize. Since reserving weights requires greater space-time cost, it is less practical than EH-FT.

\noindent
\paragraph{How does the percentage of tunable parameters in Stage 1 affect the final performance?}

There are two types of parameters to choose: original parameters (BitFit) and extra parameters (LoRA, P-Tuning). For original parameters, there is no guidance on which parameters we should train (parameters selected by BitFit are fixed). If we select randomly, the performance would be far below current PETs. EH-FT also doesn't work well in that setting. For extra parameters, we conduct experiments with EH-FT(LoRA) on RTE, BoolQ and QNLI. The result is in table \ref{ab2}.  When the middle rank $r$ increases, the score doesn't decrease significantly. This phenomena is similar to LoRA \citep{hu2021lora}. 

\begin{table}
\centering
\small{
\begin{tabular}{cccc} 
\toprule 
r & RTE  & BoolQ & QNLI \\
\midrule 
8 & \textbf{88.68} & \textbf{86.89 }& 94.73 \\
16 & 87.73 & 86.42 & 94.62 \\
32 & 87.54 & 86.30 & 94.73 \\
256 & 87.82 & 86.40 & \textbf{94.81} \\
512 & 87.20 & 86.54 & 94.75 \\
1024 & 88.26 & 86.65 & 94.79 \\
\bottomrule 
\end{tabular}
}
\caption{Results of EH-FT(LoRA) with different middle rank $r$ in Stage 1. }
\label{ab2}
\end{table}



\section{Related Work}

\textbf{Pretrained Language Models. (PLM)}  
Transformer \citep{vaswani2017attention} is a sequence-to-sequence language model with multi-head self-attention mechanism. Its encoder and decoder has become the backbone of large-scale pretrained language models like GPT \citep{radford2019language} and BERT \citep{kenton2019bert}. XLNet \citep{yang2019xlnet}, RoBERTa \citep{liu2019roberta} and DeBERTa \citep{he2020deberta} are proposed as improved models.  These models are first trained on a large corpus and then finetuned on downstream tasks, providing a significant performance gain in various NLP benchmarks. \\

\noindent
\textbf{Parameter-Efficient Tuning.}
As pretrained models continue to get larger, it becomes unacceptable to store a copy of model for each downstream task. In that case, many studies focusing on reducing trainable parameters during finetuning. Adapter \citep{houlsby2019parameter} is the first to present the concept of parameter-efficient tuning, followed by many adapter-base methods. Adapter layers are inserted between transformer layers and initialized randomly. When finetuning, all pretrained parameters are frozen and only those new 
adapter layers are trainable. Based on low intrinsic dimension of pretrained models \citep{aghajanyan2020intrinsic}, LoRA  \citep{hu2021lora} injects rank decomposition matrices into Transformer layers. Prefix Tuning  \citep{li2021prefix, liu-etal-2022-p} shows that trainable continuous prompts are also good choice. Unlike the above works, BitFit \citep{ben-zaken-etal-2022-bitfit} chose to change the bias term of the original model.\\   

\noindent 
\textbf{Generalizable Finetuning}
For large-scale pretrained models, there are many recognized problems with traditional finetuning, such as catastrophic forgetting \citep{mccloskey1989catastrophic} and overfitting \citep{jiang-etal-2020-smart}. In order to alleviate the phenomenon of catastrophic forgetting, many finetuning strategies were introduced to help model forget less. ULMFiT \citep{howard-ruder-2018-universal} proposes  triangular learning rates and gradual unfreezing. As a variant of Droupout \citep{srivastava2014dropout} , Mixout \citep{lee2020mixout} randomly mix pretrained parameters during finetuning. RecAdam \citep{chen2020recall} introduces $L_2$ distance penalty between pretrained weights and finetuned weights to prevent model weights deviating too much. Child-Tuning \citep{xu2021raise} finds that only updating a subset of parameters can obtain better generalization performance on domain transfer and task transfer. Recently, LP-FT \cite{kumar2022fine} finds that training classification head first can improve the performance for out-of-distribution data (OOD) on some computer vision benchmark. Inspired by the above approaches, EH-FT use parameter-efficient tuning to pretrain the classification head at first, which can get a better generalization ability.

\section{Conclusion}
Finetuning the pretrained model using a randomly initialized classification head in a downstream task may result in the model output feature deviating too far. We propose Efficient Head Finetuning (EH-FT), an efficient head pretraining strategy using parameter-efficient tuning and only introduce a little extra time and space while improving the model with a stable performance gain in different tasks. EH-FT can make a good initialization of head which can guide model to a local minimum close to the pretrained point, alleviating the catastrophic forgetting and overfitting of large-scale pretrained models.

Furthermore, this  method can be applied to any pretrained model, as long as it only needs a classification head for downstream tasks.
\section{Limitations}
EH-FT is an \emph{empirical method} with experiments proof currently. Although it is hard to theoretically analyze the training dynamics of large language models, it is possible to give bounds for a two-layer networks as in LP-FT, which could bring about more understanding, and we leave it for a follow-up work. 

The main measurement for feature change in this paper is based on $L_2$ distance, which, however, not the best metric. It is very possible to add a very large value in a specific dimension to increase the distance, with most value unchanged.  The method to measure the change of the feature space is also an important topic in understanding the behavior of finetuning PLMs.  



\section*{Acknowledgements}
The authors would like to thank Xiao Liu and Yue Cao for their discussion, and the reviewers of EMNLP for their valuable suggestions.

This work is supported by National Key R\&D Program of China (2021ZD0113304), National Science Foundation for Distinguished Young Scholars (No. 61825602) and Natural Science Foundation of China (No. 61836013)



\end{document}